\def\year{2018}\relax
\documentclass[letterpaper]{article} 
\usepackage{aaai19}  
\usepackage{times}  
\usepackage{helvet}  
\usepackage{courier}  
\usepackage{url}  
\usepackage{amsmath}
\usepackage{amssymb}
\usepackage{epstopdf}
\usepackage{makecell}
\usepackage{CJK}
\usepackage{url}
\usepackage{latexsym}
\usepackage{graphicx}
\usepackage{amsmath}
\usepackage{amssymb}
\usepackage{amsmath}
\usepackage{graphicx}  
\usepackage{array}
\usepackage{fp}
\usepackage{makecell}
\usepackage{flushend}
\usepackage{cases}
\usepackage{bbm}
\usepackage{soul}
\newcommand{\thickhline}{\noalign{\hrule height 1pt}}

\title{Response Generation by  Context-aware Prototype Editing}
\author{
	Yu Wu$^\dag$$^\diamondsuit$, Furu Wei$^\ddag$,	Shaohan Huang$^\ddag$~, Yunli Wang$^\dag$, Zhoujun Li$^\dag$$^\diamondsuit$, Ming Zhou$^\ddag$~~~~\\
	$^\dag$State Key Lab of Software Development Environment, Beihang University, Beijing, China\\
		$^ \diamondsuit$ Authors are supported by AdeptMind Scholarship\\
	$^\ddag$~~~~Microsoft Research, Beijing, China\\
	\{wuyu,wangyunli,lizj\}@buaa.edu.cn \{fuwei, shaohanh, mingzhou\}@microsoft.com 
}

\begin{document}
	\begin{CJK*}{UTF8}{gbsn} 
		\maketitle
		\begin{abstract}
			Open domain response generation has achieved remarkable progress in recent years, but sometimes yields short and uninformative responses. We propose a new paradigm, prototype-then-edit for response generation, that first retrieves a prototype response from a pre-defined index and then edits
			the prototype response according to the differences between the prototype context and  current context. 		
			Our motivation is that the retrieved prototype  provides a good start-point for generation because it is grammatical and informative, and the post-editing process further improves the relevance and coherence of the prototype. In practice, we design a context-aware editing model that is built upon an encoder-decoder framework augmented with an editing vector. We first generate an edit vector by considering lexical differences between a prototype context and current context. After that, the edit vector and the prototype response representation are fed to a decoder to generate a new response. Experiment results on a large scale dataset demonstrate that our new paradigm  significantly increases the  relevance, diversity and originality of generation results, compared to traditional generative models. Furthermore, our model outperforms retrieval-based methods in terms of relevance and originality. 
		\end{abstract}
		
		\section{Introduction}
		In recent years, non-task oriented chatbots focused on responding to humans intelligently on a variety of topics, have drawn much attention from both academia and industry.  Existing approaches can be categorized into generation-based methods \cite{DBLP:conf/acl/ShangLL15,vinyals2015neural,serban2015building,sordoni2015neural,serban2017hierarchical} which generate a response from scratch, and retrieval-based methods \cite{hu2014convolutional,lowe2015ubuntu,DBLP:conf/sigir/YanSW16,zhou2016multi,wu2016sequential} which select a response from an existing corpus. Since retrieval-based approaches are severely constrained by  a pre-defined index, generative approaches become more and more popular in recent years. Traditional generation-based approaches, however, do not easily generate long, diverse and informative responses, which is referred to as ``safe response" problem \cite{li2015diversity}. 
		
		%
		%
		
		\begin{table}[!t]
			
			\label{example1}  
			\centering	
			{
				\begin{tabular}{m{2cm}|m{5.5cm}}
					\hline
					Context ($c$) &  My friends and I \st{ went to some} vegan place for \st{dessert} yesterday.  \\ \thickhline
					Prototype context ($c'$)  & My friends and I \underline{had Tofu and} \underline{vegetables at a} vegan place \underline{nearby} yesterday. \\ \hline
					
					Prototype response ($r'$)  &	\textbf{Raw green vegetables} are very \textbf{beneficial} for your health.  \\ \hline
					Revised response ($r$)  & \textbf{Desserts} are very \textbf{bad}  for your health.  \\  \thickhline
					
					
				\end{tabular}
			}
			\caption{An example of context-aware prototypes editing. \underline{Underlined words} mean they do not appear in the original context, while \st{ words with strikethrough} mean they are not in the prototype context. Words in bold represent they are modified in the revised response. \label{example} }	
		\end{table}

		To address this issue, we propose a new paradigm, prototype-then-edit, for response generation.  Our motivations include: 1) human-written responses, termed as ``prototypes response", are informative, diverse and grammatical which do not suffer from short and generic issues. Hence, generating responses by editing such prototypes is able to alleviate the ``safe response" problem. 2) Some retrieved prototypes are not relevant to the current context, or suffer from a privacy issue. The post-editing process can partially solve these two problems.  3) Lexical differences between contexts provide an important signal for response editing. If a word appears in the current context but not in the prototype context, the word is likely to be inserted into the prototype response in the editing process. 

		Inspired by this idea, we formulate the response generation process as follows. Given a conversational context $c$, we first retrieve a similar context $c'$ and its associated response  $r'$ from a pre-defined index, which are called prototype context and prototype response respectively. 
		Then, we calculate an edit vector by concatenating the weighted average results of insertion word embeddings (words in prototype context but not in current context) and deletion word embeddings (words in current context but not in prototype context). 
		After that, we revise the prototype response conditioning on the edit vector. We further illustrate how our idea works with an example in  Table \ref{example}. It is obvious that the major difference between $c$ and $c'$ is what the speaker eats, so the phrase ``raw green vegetables" in $r'$ should be replaced by ``desserts" in order to adapt to the current context $c$. We hope that the decoder language model could remember the  collocation of ``desserts" and ``bad for health", so as to replace ``beneficial" with ``bad" in the revised response. The new paradigm does not only inherits the fluency and informativeness advantages from retrieval results, but also enjoys the flexibility of generation results. Hence, our edit-based model is better than previous retrieval-based and generation-based models. The edit-based model can solve the ``safe response" problem of generative models by leveraging existing responses, and is more flexible than retrieval-based models, because it does not highly depend on the index and is able to edit a response to fit current context. 
		
		
		Prior work \cite{DBLP:journals/corr/abs-1709-08878} has figured out how to edit prototype in an unconditional setting, but it cannot be applied to the response generation directly. In this paper, we propose a prototype editing method in a conditional setting\footnote{conditional setting means the editor should consider the context (dialogue history) except in a  response itself in the revision. Editor only considers a sentence itself in the unconditional setting}.  Our idea is that differences between responses strongly correlates with differences in their contexts (i.e. if a word in prototype context is changed, its related words in the response are probably modified in the editing.). 
		We realize this idea by designing a context-aware editing model that is built upon a encoder-decoder model augmented with an editing vector. The edit vector is computed by the weighted average of insertion word embeddings  and deletion word   embeddings. Larger weights mean that the editing model should pay more attention on corresponding words in revision. For instance, in Table \ref{example}, we wish words  like ``dessert", ``Tofu" and ``vegetables" get larger weights than words like ``and" and `` at". 
		The encoder learns the prototype representation with a gated recurrent unit (GRU), and feeds the representation to a decoder together with the edit vector. The decoder is a GRU language model, that regards the concatenation of last step word embedding and the edit vector as inputs, and predicts the next word with an attention mechanism.

		Our experiments are conducted on a large scale Chinese conversation corpus comprised of 20 million context-response pairs. 
		We compare our model with generative models and retrieval models in terms of fluency, relevance, diversity and originality.  The experiments show that our method outperforms traditional generative models on relevance, diversity and originality. We further find that the revised response achieves better relevance compared to its prototype and other retrieval results, demonstrating that the editing process does not only promote response originality but also improve the relevance of retrieval results. 
		
		Our contributions are listed as follows: 1) this paper proposes a new paradigm,  prototype-then-edit, for response generation; 2) we elaborate a simple but effective context-aware editing model for response generation; 3) we empirically verify the effectiveness of our method in terms of relevance, diversity, fluency and originality. 
		
		\section{Related Work}
		Research on chatbots goes back to the 1960s when ELIZA was designed \cite{weizenbaum1966eliza} with a huge amount of hand-crafted templates and rules. Recently, researchers have paid more and more attention on data-driven approaches \cite{ritter2011data,ji2014information} due to their superior scalability. Most of these methods are classified as retrieval-based methods \cite{ji2014information,DBLP:conf/sigir/YanSW16} and generation methods \cite{li2016deep,li2017adversarial,zhou2017emotional}. The former one aims to select a relevant response using a matching model, while the latter one generates a response with natural language generative models. 
		
		Prior works on retrieval-based methods mainly focus on the matching model architecture for single turn conversation \cite{hu2014convolutional} and multi-turn conversation \cite{lowe2015ubuntu,zhou2016multi,wu2016sequential}.  For the studies of generative methods, a huge amount of work aims to mitigate the ``safe response" issue from different perspectives. Most of work build models under a sequence to sequence framework \cite{sutskever2014sequence}, and introduce other elements, such as latent variables \cite{serban2017hierarchical}, topic information \cite{xing2017topic}, and dynamic vocabulary \cite{DBLP:conf/aaai/WuWYXL18} to increase response diversity.
		Furthermore, the reranking technique \cite{li2015diversity}, reinforcement learning technique \cite{li2016deep}, and adversarial learning technique \cite{li2017adversarial,DBLP:conf/emnlp/XuLWSWWQ17} have also been applied to response generation. Apart from work on ``safe response", there is a growing body of literature on style transfer \cite{DBLP:conf/aaai/FuTPZY18,DBLP:conf/emnlp/WangJBN17} and emotional response generation \cite{zhou2017emotional}. In general, most of previous work generates a response from scratch either left-to-right or conditioned on a latent vector, whereas our approach aims to generate a response by editing a prototype. 
		Prior works have attempted to utilize prototype responses to guide the generation process  \cite{song2016two,pandey2018exemplar}, in which prototype responses are encoded into vectors and feed to a decoder along with a context representation. Our work differs from previous ones on two aspects. One is they do not consider prototype context in the generation process, while our model utilizes context differences to guide editing process. The other is that we regard prototype responses as a source language, while their works formulate it as a multi-source seq2seq task, in which the current context and prototype responses are all source languages in the generation process.  
		
		Recently, some researches have explored natural language generation by editing \cite{DBLP:journals/corr/abs-1709-08878,liao2018incorporating}. A typical approach follows a writing-then-edit paradigm, that utilizes one decoder to generate a draft from scratch and uses another decoder to revise the draft \cite{DBLP:conf/nips/XiaTWLQYL17}. The other approach follows a retrieval-then-edit paradigm, that uses a Seq2Seq model to edit a  prototype retrieved from a corpus \cite{DBLP:journals/corr/abs-1709-08878,li2018delete,cao2018retrieve}. As far as we known, we are the first to leverage context lexical differences to edit prototypes.
		

		\section{Background}
		Before introducing our approach, we first briefly describe state-of-the-art natural language editing method \cite{DBLP:journals/corr/abs-1709-08878}. Given a sentence pair $(x,x')$, our goal is to obtain sentence $x$ by editing the prototype $x'$. The general framework is built upon a Seq2Seq model with an  attention mechanism, which takes $x'$ and $x$ as source sequence and target sequence respectively. The main difference is that the generative probability of a vanilla Seq2Seq model is $p(x|x')$ whereas the probability of the edit model is $p(x|x',z)$ where $z$ is an edit vector sampled from a pre-defined distribution like variational auto-encoder. In the training phase, the parameter of the distribution is conditional on the context differences. We first define $I = \{w | w\in x \wedge w\notin x' \}$ as an insertion word set, where $w$ is a word added to the prototype, and $D = \{w' | w'\in x' \wedge w'\notin x \}$ is a deletion word set, where $w$ is a word deleted from the prototype. Subsequently, we compute an insertion vector $\vec{i} = \sum_{w \in I} \Psi(w)  $ and a deletion vector $\vec{d} = \sum_{w' \in D} \Psi(w')$ by a summation over word embeddings in two corresponding sets, where $\Psi(\cdot)$ transfers a word to its embedding.  Then, the edit vector $z$ is sampled from a distribution whose parameters are governed by the concatenation of $\vec{i}$ and $ \vec{d}$. Finally, the edit vector and output of the encoder are fed to the decoder to generate $x$. 
		
		For response generation, which is a conditional setting of text editing, an interesting question raised, that is how to generate the edit by considering contexts. We will introduce our motivation and model in details in the next section.

		\section{Approach}
		\begin{figure*}[t]		
			\begin{center}
				\includegraphics[width=14cm,height=6cm]{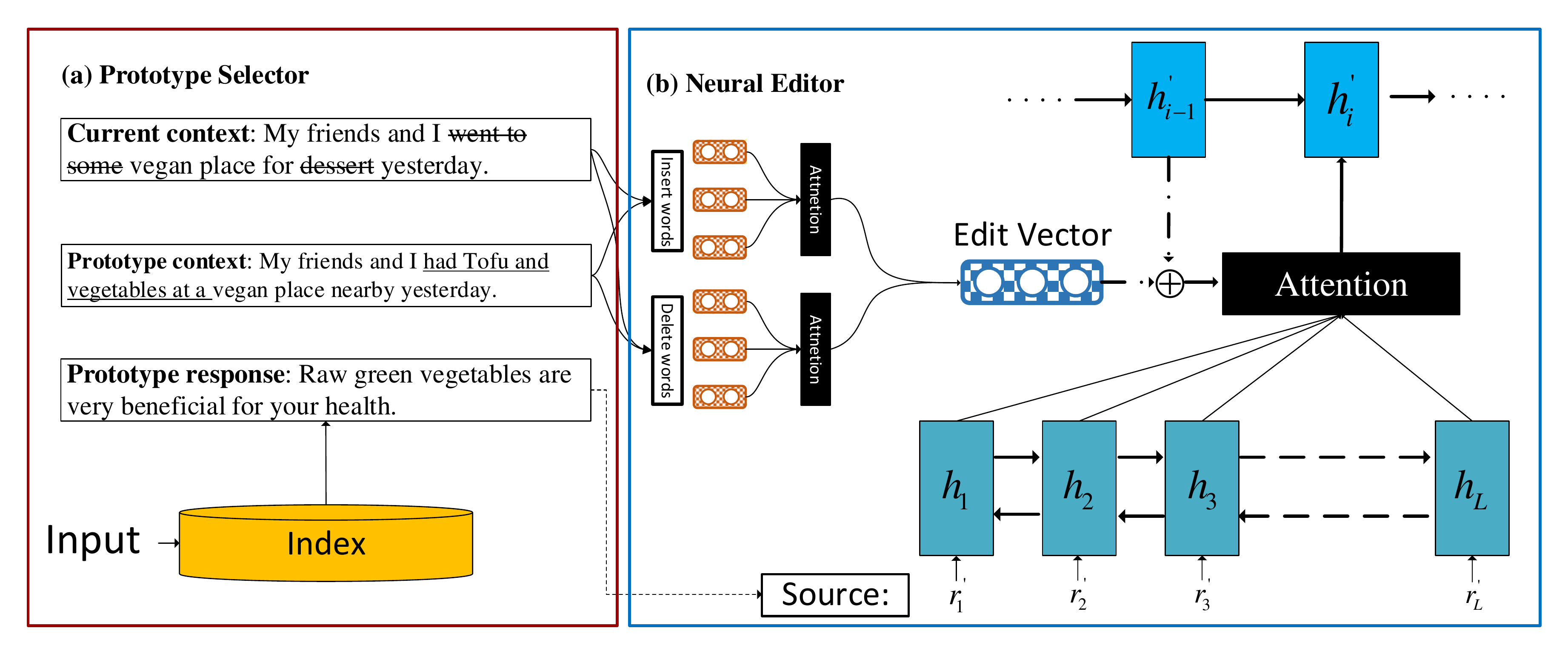}
			\end{center}
			\caption{Architecture of our model.}\label{fig:arch} 
		\end{figure*} 
		\subsection{Model Overview}
		Suppose that we have a data set $\mathcal{D} = \{(C_i,R_i)\}^N_{i=1} $. $\forall i$, $(C_i,R_i)$ comprises a context $C_i = ( c_{i,1}, \ldots, c_{i,l} ) $ and its response $R_i = ( r_{i,1}, \ldots, r_{i,l} ) $, where $c_{i,j}$ is the $j$-th word of the context $C_i$ and $r_{i,j}$ is the $j$-th word of the response $R_i$. It should be noted that $C_i$ can be either a single turn input or a multiple turn input. As the first step, we assume $C_i$ is a single turn input in this work, and leave the verification of the same technology for multi-turn response generation to future work.
		Our full model is shown in Figure \ref{fig:arch}, consisting of a prototype selector $\mathcal{S}$ and a context-aware neural editor $\mathcal{E}$.
		Given a new conversational context $C$, we first use  $\mathcal{S}$ to retrieve a context-response pair  $(C_i,R_i) \in \mathcal{D}$. Then, the editor $\mathcal{E}$  calculates an edit vector $z_i = f(C_i,C)$ to encode the information about the differences between $C_i$ and $C$. Finally, we generate a response according to the probability of $  p(R| z_i, R_i) $. In the following,   we will elaborate how to design the selector $\mathcal{S}$ and the editor $\mathcal{E}$. 
		
		\subsection{Prototype Selector} \label{selector}
		A good prototype selector $\mathcal{S}$ plays an important role in the prototype-then-edit paradigm.	We use different strategies to select prototypes for training and testing. In testing, as we described above, we retrieve a context-response pair  $(C',R')$  from a pre-defined index  for context $C$   according to the similarity of $C$ and $C'$. Here, we employ Lucene\footnote{\url{https://lucenenet.apache.org/}} to construct the index and use its inline algorithm to compute the context similarity.
		
		Now we turn to the training phase. $\forall i$, $(C_i,R_i)$, our goal is to maximize the generative probability of $R_i$ by selecting a prototype $(C'_i,R'_i)\in \mathcal{D}$.  As we already know the ground-truth response $R_i$, we first retrieve thirty prototypes $\{(C'_{i,j},R'_{i,j})\} _{j=1}^{20}$ based on the response similarity  instead of context similarity, and then reserve prototypes whose Jaccard similarity to $R_i$ are in the range of $[0.3,0.7]$. Here, we use Lucene to index all responses, and retrieve the top 20 similar responses along with their corresponding contexts for $R_i$. The Jaccard  similarity measures text similarity from a bag-of-word view, that is formulated as
		\begin{equation}
		J(A,B) = \frac{ |A \cap B| }{|A \cup B|},
		\end{equation} where $A$ and $B$ are two bags of words and $|\cdot|$ denotes the number of elements in a collection. Each context-response pair is processed with the above procedure, so we obtain enormous quadruples $\{(C_i,R_i,C'_{i,j},R'_{i,j})_{j=0}^{M_i}\}_{i=1}^N$ after this step.
		The motivation behind filtering out instances with Jaccard similarity $ < 0.3$ is that a neural editor model performs well only if a prototype is lexically similar \cite{DBLP:journals/corr/abs-1709-08878} to its ground-truth. Besides, we hope the editor does not copy the prototype so we discard instances where the prototype and groundtruth are nearly identical (i.e. Jaccard similarity $ > 0.7$). We do not use context similarity to construct parallel data for training, because similar contexts may correspond to totally different responses, so-called one-to-many phenomenon \cite{li2015diversity} in dialogue generation, that impedes editor training due to the large lexicon gap. According to our preliminary experiments, the editor always generates non-sense responses if training data is constructed by context similarity.  	


		\subsection{Context-Aware Neural Editor}
		A context-aware neural editor aims to revise a prototype to adapt current context. Formally, 
		given a quadruple $(C,R, C',R')$ (we omit subscripts  for simplification),	a context-aware neural editor first forms  an edit vector $z$ using $C$ and $C'$, and then updates parameters of the generative model by maximizing the probability of $  p(R| z, R')$. For testing, we directly generate a response after getting the editor vector. In the following, we will introduce how to obtain the edit vector and learn the generative model in details.  
		\subsubsection{Edit Vector Generation}
		For an unconditional sentence editing setting \cite{DBLP:journals/corr/abs-1709-08878}, an edit vector is randomly sampled from a distribution because how to edit the sentence is not constrained. In contrast, we should take both of $C$ and $C'$ into consideration when we revise a prototype response $R'$.  Formally, $R'$ is firstly transformed to hidden vectors $\{h_{k}| h_{k} = 	\overrightarrow{h}_{k} \oplus \overleftarrow{h}_{k} \}_{k=1}^{n_j}$ through a biGRU parameterized as Equation (\ref{biGRU}).
		\begin{equation}\label{biGRU} \small
		\overrightarrow{h}_{j}=f_{\text{GRU}} (\overrightarrow{h}_{j-1}, r'_{j}); \thickspace \overleftarrow{h}_{j}=f_{\text{GRU}} (\overleftarrow{h}_{j+1}, r'_{j})
		\end{equation} where $r'_{j}$ is the $j$-th word of $R'$.
		
		Then we compute a  context diff-vector $diff_c$ by an attention mechanism defined as follows
		\begin{equation}\label{att1}
		diff_c =  \sum_{w \in I} \beta_w \Psi(w) \oplus \sum_{w' \in D} \gamma_{w'} \Psi(w'),
		\end{equation}
		where $\oplus$ is a concatenation operation, $I = \{w | w\in C \wedge w\notin C' \}$ is a insertion word set,  and $D = \{w' | w'\in C' \wedge w'\notin C \}$ is a deletion word set. $diff_c$ explicitly encodes insertion words and deletion words from $C$ to $C'$. $\beta_w$ is the weight of a insertion word $w$, that is computed by
		\begin{eqnarray} 
		&& \beta_{w} = \frac{exp(e_{w})}{\sum_{w\in I} exp(e_{w})}, \\
		&& e_{w} = \mathbf{v_{\beta}}^\top tanh(\mathbf{W_{\beta}}[\Psi(w) \oplus h_l]),
		\end{eqnarray}
		where $ \mathbf{v_{\beta}}$ and $\mathbf{W_{\beta}}$ are parameters, and $h_l$ is the last hidden state of the encoder. $\gamma_{w'}$ is obtained with a similar process:
		\begin{eqnarray} \label{att2}
		&& \gamma_{w'} = \frac{exp(e_{w'})}{\sum_{w'\in D} exp(e_{w'})}, \\
		&& e_{w'} = \mathbf{v_{\gamma}}^\top tanh(\mathbf{W_{\gamma}}[\Psi(w') \oplus h_l]),
		\end{eqnarray}
		
		We assume that different words influence the editing process unequally, so  we weighted average insertion words and deletion words to form an edit in Equation \ref{att1}. Table \ref{example1} explains our motivation as well, that is ``desserts" is much more important than ``the" in the editing process. 
		Then we compute the edit vector $z$ by following transformation
		\begin{equation} \label{edit}
		z =  tanh(\mathbf{W}\cdot diff_c +\mathbf{b}),
		\end{equation}
		where $\mathbf{W}$ and $\mathbf{b}$ are two parameters. Equation \ref{edit} can be regarded as a mapping from context differences to response differences. 
		
		It should be noted that there are several alternative approaches to compute  $diff_c$ and $z$ for this task, such as applying memory networks, latent variables, and other complex network architectures. Here, we just use a simple method, but it yields interesting results on this task. We will further illustrate our experiment findings in the next section.
		\subsubsection{Prototype Editing}
		We build our prototype editing model upon a Seq2Seq with an attention mechanism model, which integrates the edit vector into the decoder.
		
		The decoder takes$\{h_{k} \}_{k=1}^{n_j}$ as an input and generates a response by a GRU language model with attention. The hidden state of the decoder is acquired by
		\begin{equation}\label{decoderGRU} \small
		{h}_{j}'=f_{\text{GRU}} ({h}'_{j-1}, r_{j-1} \oplus z_i),
		\end{equation} 
		where the input of $j$-th time step is the last step hidden state and the concatenation of the $(j-1)$-th word embedding and the edit vector obtained in Equation \ref{edit}.  
		Then we compute a context vector $c_i$, which is a linear combination of $\{h_1, \ldots, h_t\}$:
		\begin{equation}\small
		c_i = \sum_{j=1}^t \alpha_{i,j} h_j,
		\end{equation}
		where $\alpha_{i,j}$ is given by 
		\begin{eqnarray}\small \label{attention}
		&& \alpha_{i,j} = \frac{exp(e_{i,j})}{\sum_{k=1}^t exp(e_{i,k})}, \\
		&& e_{i,j} = \mathbf{v}^\top tanh(\mathbf{W_{\alpha}}[h_j\oplus h'_i]),
		\end{eqnarray}
		where $ \mathbf{v}$ and $\mathbf{W_{\alpha}}$ are parameters. 
		The generative probability distribution is given by
		\begin{equation} \label{attention2}
		s(r_{i} ) = softmax(\mathbf{W_p} [r_{i-1} \oplus h'_{i} \oplus c_i] +\mathbf{b_p}), 
		\end{equation}
		where $\mathbf{W_{p}}$ and $\mathbf{b_{p}}$ are two parameters. Equation \ref{attention} and \ref{attention2} are the attention mechanism \cite{bahdanau2014neural}, that mitigates the long-term dependency issue of the original Seq2Seq model. We append the edit vector to every input embedding of the decoder in Equation \ref{decoderGRU}, so the edit information can be utilized in the entire generation process. 
		
		We learn our response generation model by minimizing the negative log likelihood of $\mathcal{D}$
		\begin{equation} 
		\mathcal{L} =  -\sum_{i=1}^N \sum_{j=1}^{l} \text{log} p(r_{i,j}| z_i, R_i',r_{i,k<j}) .
		\end{equation}
		We implement our model by PyTorch \footnote{\url{https://pytorch.org/}}.
		We employ the Adam algorithm \cite{kingma2014adam} to optimize the objective function with a batch size of $128$. We set the initial learning rate as $0.001$ and reduce it by half if  perplexity on validation begins to increase. We will stop training if the perplexity on validation keeps increasing in two successive epochs. .

		\section{Experiment}
		\subsection{Experiment setting}
		In this paper, we only consider single turn response generation. We collected over
		20 million human-human context-response pairs (context only contains 1 turn) from Douban Group \footnote{\url{https://www.douban.com/group} Douban is a popular forum in China.}. After removing	duplicated pairs and utterance longer than	30 words, we split 19,623,374 pairs for training, 	10,000 pairs for validation and 10,000 pairs for	testing. The average length of contexts and responses	are 11.64 and 12.33 respectively. The training data mentioned above is used by retrieval models and generative models. 
		
		In terms of ensemble models and our editing model, the validation set and the test set are the same with  datasets prepared for retrieval and generation models. Besides, for each context in the validation and test sets, we select its prototypes  with the method described in Section ``Prototype Selector". We follow Song et al. \shortcite{song2016two} to construct a training data set for ensemble models, and construct a training data set with the method described in Section ``Prototype Selector" for our editing models. We can obtain 42,690,275 $(C_i,R_i,C'_i,R'_i)$ quadruples with the proposed data preparing method. For a fair comparison,	we randomly sample 19,623,374 instances for the training of our method and the ensemble method respectively. To facilitate further research, related resources of the paper can be found at \url{https://github.com/MarkWuNLP/ResponseEdit}. 

		\subsection{Baseline}
		%
		%
		%
		%
		%
		
		\begin{table*}[!t]
			\small
			\centering
			
			\caption{Automatic evaluation results. Numbers in bold mean that improvement from the model on that metric is statistically significant over the baseline methods (t-test, p-value $<0.01$).  $\kappa$ denotes Fleiss Kappa \cite{fleiss1971measuring}, which reflects the agreement among human annotators. 
			}		\label{exp} 	
			\scalebox{0.95}{
			\begin{tabular}{l|c|c|c|c|c|c|c|c|c|c|c}
				\hline
				&   \multicolumn{3}{c|}{\textbf{Relevance}} 	&   \multicolumn{2}{c|}{\textbf{Diversity}} & \multicolumn{1}{c|}{\textbf{Originality}} & \multicolumn{5}{c}{\textbf{Fluency}}\\\hline
				&  Average&  Extrema &  Greedy &  Distinct-1 &  Distinct-2 & Not appear & +2 & +1 & 0 & Avg & $\kappa$\\ \hline
				
				S2SA &0.346&0.180&0.350&0.032& 0.087 &0.208 & 94.0\% & 5.2\% & 0.8\%& \textbf{1.932} &0.89\\ 
				S2SA-MMI&0.379&0.189&0.385&0.039& 0.127 & 0.297 & 91.6\% & 7.6\% & 0.8\%& 1.908&0.83\\ 	 		
				CVAE &0.360&0.183&0.363& 0.062&0.178 & 0.745&  83.8\% & 12.7\% & 3.5\%&1.803&0.84\\ \hline
				Retrieval-default &0.288&0.130&0.309&0.098&\textbf{0.589} &0.000 & 92.8\%  & 6.8\%  & 0.4\% &1.924 &0.88\\ 
				Retrieval-Rerank& 0.380&0.191&0.381&\textbf{0.106}&0.560 &0.000 & 91.7\%  & 7.8\%  & 0.5\% &1.912 &0.88\\ 
				Ensemble-default &0.352&0.183&0.362&0.035&0.097 &0.223 & 91.3\%  & 7.2\%  & 1.5\% &1.898 &0.84\\ 
				Ensemble-Rerank&0.372&0.187&0.379&0.040&0.135 &0.275 & 90.4\%  & 7.8\%  & 1.8\% &1.898 &0.84\\ \hline
				Edit-default&0.297&0.150&0.327& 0.071&0.300&0.796 & 89.6\% & 9.2\% & 1.2\%&1.884&0.87 \\ 	
				Edit-1-Rerank& 0.367&0.185&0.371&0.077&0.296&0.794 & 93.2\% & 5.6\% & 1.2\%&1.920&0.79 \\
				Edit-N-Rerank& \textbf{0.386}& \textbf{0.203}& \textbf{0.389}&0.068&0.280 & \textbf{0.860} & 87.2\% & 10.8\% & 2.0\%&1.852&0.85\\  \hline
				Ensemble-Merge&0.385&0.193&0.387& 0.058&0.247&0.134 & 91.2\% & 7.3\% & 1.5\%&1.897&0.89 \\ 	
				Edit-Merge& 0.391&0.197&0.392&0.103&0.443&0.734 & 88.6\% & 9.7\% & 1.7\%&1.869&0.85 \\
				\hline
			\end{tabular}	
	}
		\end{table*}
		
		\textbf{S2SA}: We apply the Seq2Seq with attention \cite{bahdanau2014neural} as a baseline model. We use a Pytorch implementation, OpenNMT  \cite{opennmt} in the experiment.  
		
		\textbf{S2SA-MMI}: We employed the bidirectional-MMI decoder as in \cite{li2015diversity}. The hyperparameter $\lambda$ is set as 0.5 according to the paper's suggestion. 200 candidates are sampled from beam search for reranking.

		\textbf{CVAE}: The conditional variational auto-encoder is a popular method of increasing the diversity of response generation \cite{DBLP:conf/acl/ZhaoZE17}. We use the published code at \url{https://github.com/snakeztc/NeuralDialog-CVAE}, and conduct small adaptations for our single turn scenario.

		\textbf{Retrieval}: We compare our model with two retrieval-based methods to show the effect of editing. One is \textbf{Retrieval-default} that  directly regards the top-1 result given by Lucene as the reply. The second one is \textbf{Retrieval-Rerank}, where we first retrieve 20 response candidates, and then employ a dual-LSTM model \cite{lowe2015ubuntu} to compute matching degree between current context and the candidates. The matching model is implemented  with the same setting in \cite{lowe2015ubuntu}, and is trained on the training data set where negative instances are randomly sampled with a ratio of $pos:neg = 1:9$. 
		
		\textbf{Ensemble Model}: Song et al \shortcite{song2016two} propose an ensemble of retrieval and generation methods.  It encodes current context and retrieved responses (Top-k retrieved responses are all used in the generation process.) into vectors, and feeds these representations to a decoder to generate a new response. As there is no official code, we implement it carefully by ourselves. We use the top-1 response returned by beam search as a baseline, denoted as \textbf{Ensemble-default}. For a fair comparison, we further rerank top 20 generated results with the same LSTM based matching model, and denote it as \textbf{Ensemble-Rerank}. We further create a candidate pool by merging the retrieval and generation results, and rerank them with the same ranker. The method is denoted as \textbf{Ensemble-Merge}.
		
		Correspondingly, we evaluate three variants of our model. Specifically,  \textbf{Edit-default} and \textbf{Edit-1-Rerank}  edit top-1 response yielded by Retrieval-default and Retrieval-Rerank respectively.  \textbf{Edit-N-Rerank} edits all 20 responses returned by Lucene and then reranks the revised results with the dual-LSTM model. We also merge edit results of \textbf{Edit-N-Rerank} and candidates returned by the search engine, and then rerank them, which is denoted as \textbf{Edit-Merge}. 
		In practice, the word embedding size and editor vector size are 512, and both of the encoder and decoder are a 1-layer GRU whose hidden vector size is 1024. Message and response vocabulary size are 30000, and words not covered by the vocabulary are represented by a placeholder \$UNK\$. Word embedding size, hidden vector size and attention vector size of baselines and our models are the same. All generative models use beam search to yield responses, where the beam size is 20 except S2SA-MMI. For all models, we remove \$UNK\$ from the target vocabulary, because it always leads to a fluency issue in evaluation.

		\subsection{Evaluation Metrics}
		We evaluate our model on four criteria: fluency, relevance, diversity and originality. 	We employ Embedding Average (Average), Embedding Extrema (Extrema), and Embedding Greedy (Greedy) \cite{liu2016not} to evaluate response relevance, which are better correlated with human judgment than BLEU. Following \cite{li2015diversity}, we evaluate the response diversity based on the ratios of distinct unigrams and bigrams in generated responses, denoted as Distinct-1 and Distinct-2. In this paper, we define a new metric, originality, that is defined as the ratio of generated responses that do not appear in the training set. Here, ``appear" means we can find exactly the same response in our training data set. 	We randomly select 1,000 contexts from the test set, and ask three native speakers to annotate response fluency. We conduct 3-scale rating: +2, +1 and 0. +2: The response is fluent and grammatically correct. +1: There are a few grammatical errors in the response but readers could understand it. 0: The response is totally grammatically broken, making it difficult to understand. As how to evaluate response generation automatically is still an open problem \cite{liu2016not}, we further conduct human evaluations to compare our models with baselines. We ask the same three native speakers to do a side-by-side comparison \cite{li2016deep} on the 1,000 contexts.  Given a context and two responses generated by different models, we ask annotators to decide which response is better (Ties are permitted). 

		\begin{table}[h]
			\centering
			\caption{Human side-by-side evaluation results.   If a row is labeled as ``a v.s.b", the second column, ``loss", means the ratio of responses given by ``a'' are worse than those given by``b''.  
			}		\label{exp:human} 	
			\scalebox{0.9}{
			\begin{tabular}{l|c|c|c|c}
				\hline
				& Loss & Tie & Win & $\kappa$ \\ \hline
				
				Ed-Default v.s. R-Default &23.6&41.2&35.2 & 0.54\\
				Ed-Default v.s. Ens-Default &29.4&47.8& 22.8 &0.59 \\\hline
				Ed-1-Rerank v.s. R-Rerank&29.2&45.3&25.5 & 0.60\\
				Ed-N-Rerank v.s. Ens-Rerank &24.9&42.1&33.0 & 0.55\\
				Ed-N-Rerank v.s. R-Rerank &25.2&44.8&30.0 & 0.62\\ \hline
				
				\hline
			\end{tabular}	}
		\end{table}
		
		%
		%
		%
		%
		%
		%
		%
		\subsection{Evaluation Results}
		Table \ref{exp} shows the evaluation results on the Chinese dataset. Our methods are better than retrieval-based methods on embedding based metrics, that means revised responses are more relevant to ground-truth in the semantic space. Our model just slightly revises prototype response, so improvements on automatic metrics are not that large but significant on statistical tests (t-test, p-value $<0.01$). Two factors are known to cause Edit-1-Rerank worse than Retrieval-Rerank. 1) Rerank algorithm is biased to long responses, that poses a challenge for the editing model. 2) Despite of better prototype responses, a context of top-1 response is always greatly different from current context, leading to  a large insertion word set and a large deletion set, that also obstructs the revision process. In terms of diversity, our methods drop on distinct-1 and distinct-2 in a comparison with retrieval-based methods, because the editing model often deletes special words pursuing for better relevance. Retrieval-Rerank is better than retrieval-default, indicating that it is necessary to rerank responses by measuring context-response similarity with a matching model.  
		
		Our methods significantly outperform generative baselines in terms of diversity since prototype  responses are good start-points that are diverse and informative. It demonstrates that the prototype-then-editing paradigm is capable of addressing the safe response problem.  Edit-Rerank is better than generative baselines on relevance but Edit-default is not, indicating a good prototype selector is quite important to our editing model. In terms of originality, about 86$\%$ revised response do not appear in the training set, that surpasses S2SA, S2SA-MMI and CVAE. This is mainly because baseline methods are more likely to generate safe responses that are frequently appeared in the training data, while our model tends to modify an existing response that avoids duplication issue. In terms of fluency, S2SA achieves the best results, and retrieval based approaches come to the second place. Safe response enjoys high score on fluency, that is why S2SA and S2SA-MMI perform well on this metric. Although editing based methods are not the best on the fluency metric, they also achieve a high absolute number. That is an acceptable fluency score for a dialogue engine, indicating that most of generation responses are grammatically correct. In addition, in terms of the fluency metric, Fleiss' Kappa \cite{fleiss1971measuring} on all models are around 0.8, showing a high agreement among labelers.
%
%
%

		Compared to ensemble models, our model performs much better on diversity and originality, that is because we regard prototype response instead of the current context as source sentence in the Seq2Seq, which keeps most of content in prototype but slightly revises it based on the context difference. Both of the ensemble and edit model are improved when the original retrieval candidates are considered in the rerank process.
		
		Regarding  human side-by-side evaluation, we can find that Edit-Default and Edit-N-rerank are slightly better than Retrieval-default and Retrieval-rerank (The winning examples are more than the losing examples), indicating that the post-editing is able to improve the response quality. Ed-Default is worse than Ens-Default, but Ed-N-Rerank is better than Ens-Rerank. This is mainly because the editing model regards the prototype response as the source language, so it is highly depends on the quality of prototype response. 

		
		%
		%
		

		\begin{table}[h]
				\small
			\centering
			\caption{Model ablation tests. Full model denotes Edit-N-Rerank. ``-del" means we only consider insertion words. ``-ins" means we only consider deletion words. ``-both" means we train a standard seq2seq model from prototype response to revised response.
			}		\label{exp:ablation} 	
		
			\scalebox{0.8}{
				\begin{tabular}{l|c|c|c|c|c|c}
					\hline
					& Average & Extrema & Greedy & Dis-1 & Dis-2& Originality \\ \hline
					
					Full &38.6&20.3&38.9 &0.068 & 0.280 & 86.0\\
					- ins &34.2&17.9&34.8 &0.067 &0.286 & 85.1\\
					- del  &34.0&17.6&34.6 &0.065 &0.278 &  86.3 \\
					-both  &32.8&16.9&33.7 &0.067 &0.282 &  85.4 \\
					\hline
			\end{tabular}		}
		\end{table}
	\begin{table}[!t]
		\small
		\centering	
		{	\caption{Case Study. We show examples yielded by Edit-default and Edit-Rerank. Chinese utterances are translated to English here.  \label{case study}}	
			\begin{tabular}{m{1.5cm}|m{5.5cm}}
				\hline
				& Insertion case\\ \hline
				Context &  {身}~ 在 ~国外~ 寂寞 ~无聊 ~就 ~化妆
				\\ & When I feel bored and lonely abroad, I like makeup.
				\\ \hline
				Prototype context & 无聊~ 就~ 玩\\&
				If I feel bored, I will play something.
				\\ & 
				\\ \hline
				Prototype response &	嗯  \\ & OK. 
				\\ \hline
				Revised response & 嗯 ~我~ 也~ 喜欢~化妆
				\\ &  OK. I love makeup too. 
				\\  \thickhline
				& Deletion case\\ \hline
				
				Context &  我~ {比较}\color{black}~ 常 ~吃~ 辛拉面 \\ & I often eat spice noodles. \\\hline
				Prototype context & 我~在~台湾~时候~常~吃~辛拉面~和~卤肉饭。 \\ & When I lived in Taiwan, I often eat spicy Noodles and braised pork rice.  \\ \hline
				
				Prototype response &	我~也~喜欢~卤肉饭。
				\\ & I love braised pork rice as well. 
				\\ \hline
				Revised response & 我~也~喜欢。 \\ & I love it as well. (In Chinese, model just deletes the phrase ``braised pork rice" without adding any word. )	\\	\hline			  
				
				& Replacement case\\ \hline
				Context &  纹身~有没有~办法~全部~弄~干净 \\ & Is there any way to get all tattoos clean\\\hline
				Prototype context &  把~药~抹~头发~上~能~把~头发~弄~干净 ~么 \\ & Is it possible to clean your hair by wiping the medicine on your hair? \\ \hline
				
				Prototype response &	抹~完~真的~头发~干净~很多  \\
				&After wiping it, hair gets clean much.
				\\ \hline
				Revised response & 抹完~纹身~就~ 会~ 掉~ 很多 \\
				&After wiping it, most of tattoos will be cleaned. 
				\\  \thickhline
			\end{tabular}
		}
		
	\end{table}

		\subsection{Discussions}
		\subsubsection{Model ablation} 
		We train variants of our model by removing the insertion word vector, the deletion word vector, and both of them respectively. The results are shown in Table \ref{exp:ablation}. We can find that embedding based metrics drop dramatically when the editing vector is partially or totally removed, indicating that the edit vector is crucial for response relevance. Diversity and originality do not decrease after the edit vector is removed, implying that the retrieved prototype is the key factor for these two metrics. According to above observations, we conclude that the prototype selector and the context-aware editor play different roles in generating responses. 
		\subsubsection{Editing Type Analysis} It is interesting to explore the semantic gap between prototype and revised response. We ask annotators to conduct 4-scale rating on 500 randomly sampled prototype-response pairs given by Edit-default and Edit-N-Rerank respectively. The 4-scale is defined as:  identical, paraphrase, on the same topic and unrelated.
		
		\begin{figure}[h]		
			\begin{center}
				\includegraphics[width=7.5cm,height=4.5cm]{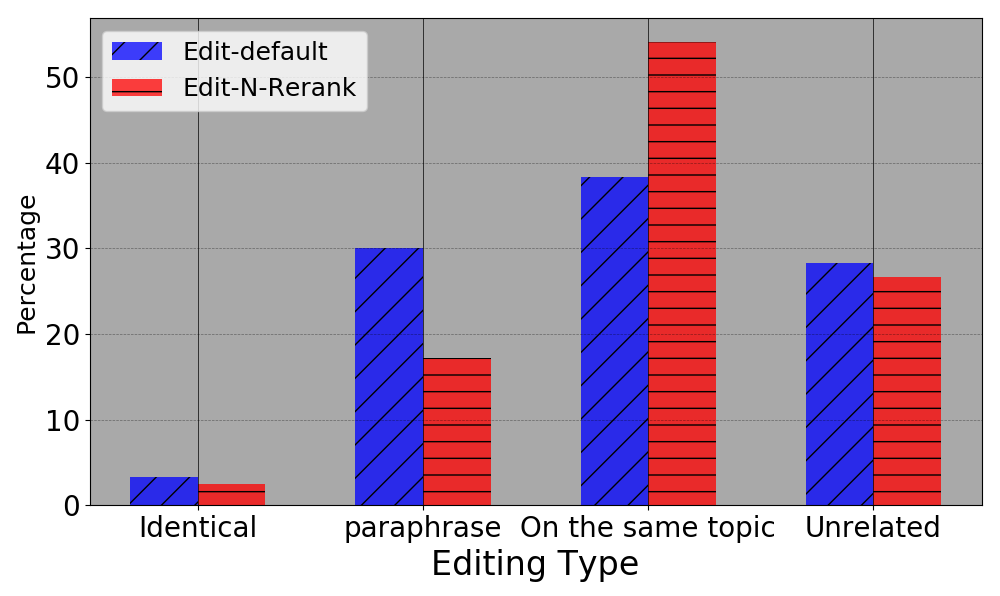}
			\end{center}	
			\caption{Distribution across different editing types.}\label{fig:ratio} 
			
		\end{figure}
		Figure \ref{fig:ratio} provides the ratio of four editing types defined above. For both methods, Only $2\% $  of edits are exactly the same with the prototype, that means our model does not downgrade to a copy model. Surprisingly, there are $30\%$ revised responses are unrelated to prototypes. The key factor for this phenomenon is that the neural editor will rewrite the prototype when it is hard to insert insertion words to the prototype.   The ratio of ``on the same topic" response given by Edit-N-rerank is larger than Edit-default, revealing that ``on the same topic" responses might be more relevant from the view of a LSTM based reranker. 
		\subsubsection{Case Study} 
		
%
%
%
%
		We give three examples to show how our model works in Table \ref{case study}. The first case illustrates the effect of word insertion. Our editing model enriches a short response by inserting words from context, that makes the conversation informative and coherent. The second case gives an example of word deletion, where a phrase ``braised pork rice" is removed as it does not fit current context. Phrase ``braised pork rice" only appears in the prototype context but not in current context, so it is in the deletion word set $D$, that makes the decoder not generate it.  
	The third one is that our model  forms a relevant query by deleting some words in the prototype while inserting other words to it.  Current context is talking about ``clean tatoo", but the prototype discusses ``clean hair", leading to an irrelevant response. After the word substitution, the revised response becomes appropriated for current context.
	 \subsubsection{Editing Words} According to our observation, function words and nouns are more likely to be added/deleted. This is mainly because function words, such as pronoun, auxiliary, and interjection may be substituted in the paraphrasing. In addition, a large proportion of context differences is caused by nouns substitutions, thus we observe that nouns are added/deleted in the revision frequently.

		
		\section{Conclusion}
		We present a new paradigm, prototype-then-edit, for open domain response generation, that  enables a generation-based chatbot to leverage retrieved results. We propose a simple but effective model to edit context-aware responses by taking context differences into consideration.   
		Experiment results on a large-scale dataset show that our model outperforms traditional methods on some metrics.  In the future, we will investigate how to jointly learn the prototype selector and neural editor.
		
\section{Acknowledgments}
Yu is supported by AdeptMind Scholarship and Microsoft Scholarship. This work was supported in part by the Natural Science Foundation of China (Grand Nos. U1636211, 61672081, 61370126), Beijing Advanced Innovation Center for Imaging Technology (No.BAICIT-2016001) and National Key R\&D Program of China (No.2016QY04W0802). 
	\bibliography{acl2018}

\begin{thebibliography}{}

\bibitem[\protect\citeauthoryear{Bahdanau, Cho, and
  Bengio}{2015}]{bahdanau2014neural}
Bahdanau, D.; Cho, K.; and Bengio, Y.
\newblock 2015.
\newblock Neural machine translation by jointly learning to align and
  translate.
\newblock {\em ICLR}.

\bibitem[\protect\citeauthoryear{Cao \bgroup et al\mbox.\egroup
  }{2018}]{cao2018retrieve}
Cao, S.~Z.; Li, W.; Wei, F.; and Li, S.
\newblock 2018.
\newblock Retrieve, rerank and rewrite: Soft template based neural
  summarization.
\newblock In {\em ACL}, volume~2, ~3.

\bibitem[\protect\citeauthoryear{Fleiss}{1971}]{fleiss1971measuring}
Fleiss, J.~L.
\newblock 1971.
\newblock Measuring nominal scale agreement among many raters.
\newblock {\em Psychological bulletin} 76(5):378.

\bibitem[\protect\citeauthoryear{Fu \bgroup et al\mbox.\egroup
  }{2018}]{DBLP:conf/aaai/FuTPZY18}
Fu, Z.; Tan, X.; Peng, N.; Zhao, D.; and Yan, R.
\newblock 2018.
\newblock Style transfer in text: Exploration and evaluation.
\newblock In {\em AAAI 2018}.

\bibitem[\protect\citeauthoryear{Guu \bgroup et al\mbox.\egroup
  }{2017}]{DBLP:journals/corr/abs-1709-08878}
Guu, K.; Hashimoto, T.~B.; Oren, Y.; and Liang, P.
\newblock 2017.
\newblock Generating sentences by editing prototypes.
\newblock {\em CoRR} abs/1709.08878.

\bibitem[\protect\citeauthoryear{Hu \bgroup et al\mbox.\egroup
  }{2014}]{hu2014convolutional}
Hu, B.; Lu, Z.; Li, H.; and Chen, Q.
\newblock 2014.
\newblock Convolutional neural network architectures for matching natural
  language sentences.
\newblock In {\em NIPS},  2042--2050.

\bibitem[\protect\citeauthoryear{Ji, Lu, and Li}{2014}]{ji2014information}
Ji, Z.; Lu, Z.; and Li, H.
\newblock 2014.
\newblock An information retrieval approach to short text conversation.
\newblock {\em arXiv preprint arXiv:1408.6988}.

\bibitem[\protect\citeauthoryear{Kingma and Ba}{2015}]{kingma2014adam}
Kingma, D., and Ba, J.
\newblock 2015.
\newblock Adam: A method for stochastic optimization.
\newblock {\em ICLR}.

\bibitem[\protect\citeauthoryear{Klein \bgroup et al\mbox.\egroup
  }{2017}]{opennmt}
Klein, G.; Kim, Y.; Deng, Y.; Senellart, J.; and Rush, A.~M.
\newblock 2017.
\newblock Opennmt: Open-source toolkit for neural machine translation.
\newblock In {\em Proc. ACL}.

\bibitem[\protect\citeauthoryear{Li \bgroup et al\mbox.\egroup
  }{2016a}]{li2015diversity}
Li, J.; Galley, M.; Brockett, C.; Gao, J.; and Dolan, B.
\newblock 2016a.
\newblock A diversity-promoting objective function for neural conversation
  models.
\newblock In {\em NAACL, 2016},  110--119.

\bibitem[\protect\citeauthoryear{Li \bgroup et al\mbox.\egroup
  }{2016b}]{li2016deep}
Li, J.; Monroe, W.; Ritter, A.; Jurafsky, D.; Galley, M.; and Gao, J.
\newblock 2016b.
\newblock Deep reinforcement learning for dialogue generation.
\newblock In {\em EMNLP 2016},  1192--1202.

\bibitem[\protect\citeauthoryear{Li \bgroup et al\mbox.\egroup
  }{2017}]{li2017adversarial}
Li, J.; Monroe, W.; Shi, T.; Ritter, A.; and Jurafsky, D.
\newblock 2017.
\newblock Adversarial learning for neural dialogue generation.
\newblock {\em EMNLP}.

\bibitem[\protect\citeauthoryear{Li \bgroup et al\mbox.\egroup
  }{2018}]{li2018delete}
Li, J.; Jia, R.; He, H.; and Liang, P.
\newblock 2018.
\newblock Delete, retrieve, generate: A simple approach to sentiment and style
  transfer.
\newblock {\em NAACL}.

\bibitem[\protect\citeauthoryear{Liao \bgroup et al\mbox.\egroup
  }{2018}]{liao2018incorporating}
Liao, Y.; Bing, L.; Li, P.; Shi, S.; Lam, W.; and Zhang, T.
\newblock 2018.
\newblock Incorporating pseudo-parallel data for quantifiable sequence editing.
\newblock {\em arXiv preprint arXiv:1804.07007}.

\bibitem[\protect\citeauthoryear{Liu \bgroup et al\mbox.\egroup
  }{2016}]{liu2016not}
Liu, C.-W.; Lowe, R.; Serban, I.~V.; Noseworthy, M.; Charlin, L.; and Pineau,
  J.
\newblock 2016.
\newblock How not to evaluate your dialogue system: An empirical study of
  unsupervised evaluation metrics for dialogue response generation.
\newblock {\em EMNLP}.

\bibitem[\protect\citeauthoryear{Lowe \bgroup et al\mbox.\egroup
  }{2015}]{lowe2015ubuntu}
Lowe, R.; Pow, N.; Serban, I.; and Pineau, J.
\newblock 2015.
\newblock The ubuntu dialogue corpus: A large dataset for research in
  unstructured multi-turn dialogue systems.
\newblock {\em SIGDIAL}.

\bibitem[\protect\citeauthoryear{Pandey \bgroup et al\mbox.\egroup
  }{2018}]{pandey2018exemplar}
Pandey, G.; Contractor, D.; Kumar, V.; and Joshi, S.
\newblock 2018.
\newblock Exemplar encoder-decoder for neural conversation generation.
\newblock In {\em ACL}, volume~1,  1329--1338.

\bibitem[\protect\citeauthoryear{Ritter, Cherry, and
  Dolan}{2011}]{ritter2011data}
Ritter, A.; Cherry, C.; and Dolan, W.~B.
\newblock 2011.
\newblock Data-driven response generation in social media.
\newblock In {\em EMNLP},  583--593.

\bibitem[\protect\citeauthoryear{Serban \bgroup et al\mbox.\egroup
  }{2016}]{serban2015building}
Serban, I.~V.; Sordoni, A.; Bengio, Y.; Courville, A.~C.; and Pineau, J.
\newblock 2016.
\newblock End-to-end dialogue systems using generative hierarchical neural
  network models.
\newblock In {\em Proceedings of the Thirtieth {AAAI} Conference on Artificial
  Intelligence, February 12-17, 2016, Phoenix, Arizona, {USA.}},  3776--3784.

\bibitem[\protect\citeauthoryear{Serban \bgroup et al\mbox.\egroup
  }{2017}]{serban2017hierarchical}
Serban, I.~V.; Sordoni, A.; Lowe, R.; Charlin, L.; Pineau, J.; Courville,
  A.~C.; and Bengio, Y.
\newblock 2017.
\newblock A hierarchical latent variable encoder-decoder model for generating
  dialogues.
\newblock In {\em AAAI},  3295--3301.

\bibitem[\protect\citeauthoryear{Shang, Lu, and
  Li}{2015}]{DBLP:conf/acl/ShangLL15}
Shang, L.; Lu, Z.; and Li, H.
\newblock 2015.
\newblock Neural responding machine for short-text conversation.
\newblock In {\em {ACL} 2015}.

\bibitem[\protect\citeauthoryear{Song \bgroup et al\mbox.\egroup
  }{2016}]{song2016two}
Song, Y.; Yan, R.; Li, X.; Zhao, D.; and Zhang, M.
\newblock 2016.
\newblock Two are better than one: An ensemble of retrieval-and
  generation-based dialog systems.
\newblock {\em arXiv preprint arXiv:1610.07149}.

\bibitem[\protect\citeauthoryear{Sordoni \bgroup et al\mbox.\egroup
  }{2015}]{sordoni2015neural}
Sordoni, A.; Galley, M.; Auli, M.; Brockett, C.; Ji, Y.; Mitchell, M.; Nie, J.;
  Gao, J.; and Dolan, B.
\newblock 2015.
\newblock A neural network approach to context-sensitive generation of
  conversational responses.
\newblock In {\em NAACL 2015},  196--205.

\bibitem[\protect\citeauthoryear{Sutskever, Vinyals, and
  Le}{2014}]{sutskever2014sequence}
Sutskever, I.; Vinyals, O.; and Le, Q.~V.
\newblock 2014.
\newblock Sequence to sequence learning with neural networks.
\newblock In {\em Advances in neural information processing systems},
  3104--3112.

\bibitem[\protect\citeauthoryear{Vinyals and Le}{2015}]{vinyals2015neural}
Vinyals, O., and Le, Q.
\newblock 2015.
\newblock A neural conversational model.
\newblock {\em arXiv preprint arXiv:1506.05869}.

\bibitem[\protect\citeauthoryear{Wang \bgroup et al\mbox.\egroup
  }{2017}]{DBLP:conf/emnlp/WangJBN17}
Wang, D.; Jojic, N.; Brockett, C.; and Nyberg, E.
\newblock 2017.
\newblock Steering output style and topic in neural response generation.
\newblock In {\em EMNLP 2017},  2140--2150.

\bibitem[\protect\citeauthoryear{Weizenbaum}{1966}]{weizenbaum1966eliza}
Weizenbaum, J.
\newblock 1966.
\newblock Eliza: a computer program for the study of natural language
  communication between man and machine.
\newblock {\em Communications of the ACM} 9(1):36--45.

\bibitem[\protect\citeauthoryear{Wu \bgroup et al\mbox.\egroup
  }{2017}]{wu2016sequential}
Wu, Y.; Wu, W.; Xing, C.; Zhou, M.; and Li, Z.
\newblock 2017.
\newblock Sequential matching network: {A} new architecture for multi-turn
  response selection in retrieval-based chatbots.
\newblock In {\em ACL 2017},  496--505.

\bibitem[\protect\citeauthoryear{Wu \bgroup et al\mbox.\egroup
  }{2018}]{DBLP:conf/aaai/WuWYXL18}
Wu, Y.; Wu, W.; Yang, D.; Xu, C.; and Li, Z.
\newblock 2018.
\newblock Neural response generation with dynamic vocabularies.
\newblock In {\em AAAI 2018}.

\bibitem[\protect\citeauthoryear{Xia \bgroup et al\mbox.\egroup
  }{2017}]{DBLP:conf/nips/XiaTWLQYL17}
Xia, Y.; Tian, F.; Wu, L.; Lin, J.; Qin, T.; Yu, N.; and Liu, T.
\newblock 2017.
\newblock Deliberation networks: Sequence generation beyond one-pass decoding.
\newblock In {\em NIPS 2017},  1782--1792.

\bibitem[\protect\citeauthoryear{Xing \bgroup et al\mbox.\egroup
  }{2017}]{xing2017topic}
Xing, C.; Wu, W.; Wu, Y.; Liu, J.; Huang, Y.; Zhou, M.; and Ma, W.-Y.
\newblock 2017.
\newblock Topic aware neural response generation.
\newblock In {\em AAAI}, volume~17,  3351--3357.

\bibitem[\protect\citeauthoryear{Xu \bgroup et al\mbox.\egroup
  }{2017}]{DBLP:conf/emnlp/XuLWSWWQ17}
Xu, Z.; Liu, B.; Wang, B.; Sun, C.; Wang, X.; Wang, Z.; and Qi, C.
\newblock 2017.
\newblock Neural response generation via {GAN} with an approximate embedding
  layer.
\newblock In {\em EMNLP 2017},  617--626.

\bibitem[\protect\citeauthoryear{Yan, Song, and
  Wu}{2016}]{DBLP:conf/sigir/YanSW16}
Yan, R.; Song, Y.; and Wu, H.
\newblock 2016.
\newblock Learning to respond with deep neural networks for retrieval-based
  human-computer conversation system.
\newblock In {\em SIGIR 2016},  55--64.

\bibitem[\protect\citeauthoryear{Zhao, Zhao, and
  Esk{\'{e}}nazi}{2017}]{DBLP:conf/acl/ZhaoZE17}
Zhao, T.; Zhao, R.; and Esk{\'{e}}nazi, M.
\newblock 2017.
\newblock Learning discourse-level diversity for neural dialog models using
  conditional variational autoencoders.
\newblock In {\em ACL 2017},  654--664.

\bibitem[\protect\citeauthoryear{Zhou \bgroup et al\mbox.\egroup
  }{2016}]{zhou2016multi}
Zhou, X.; Dong, D.; Wu, H.; Zhao, S.; Yu, D.; Tian, H.; Liu, X.; and Yan, R.
\newblock 2016.
\newblock Multi-view response selection for human-computer conversation.
\newblock In {\em EMNLP 2016},  372--381.

\bibitem[\protect\citeauthoryear{Zhou \bgroup et al\mbox.\egroup
  }{2017}]{zhou2017emotional}
Zhou, H.; Huang, M.; Zhang, T.; Zhu, X.; and Liu, B.
\newblock 2017.
\newblock Emotional chatting machine: Emotional conversation generation with
  internal and external memory.
\newblock {\em arXiv preprint arXiv:1704.01074}.

\end{thebibliography}
	\bibliographystyle{aaai}
	\end{CJK*}
\end{document}